\documentclass[10pt,twocolumn,letterpaper]{article}

\usepackage{iccv}
\usepackage{times}
\usepackage{epsfig}
\usepackage{graphicx}
\usepackage{amsmath}
\usepackage{amssymb}
\usepackage{hhline}
\usepackage{subcaption}
\usepackage{mathtools, nccmath}
\usepackage{multirow}

\usepackage{enumitem}

\usepackage[pagebackref=true,breaklinks=true,letterpaper=true,colorlinks,bookmarks=false]{hyperref}

\iccvfinalcopy

\ificcvfinal\fi

\begin{document}

\title{Controllable Video Generation by Learning the Underlying Dynamical System with Neural~ODE}

\author{Yucheng Xu$^1$, Li Nanbo$^1$, Arushi Goel$^1$, Zijian Guo$^3$, Zonghai Yao$^4$ \\ Hamidreza Kasaei$^2$, Mohammadreze Kasaei$^1$, Zhibin Li$^{1,5}$\\
University of Edinburgh$^1$, University of Groningen$^2$, Carnegie Mellon University$^3$ \\ University of Massachusetts Amherst$^4$, University College London$^5$\\
{\tt\small \{yucheng.xu, nanbo.li, m.kasaei, a.goel-1\}@ed.ac.uk$^1$}\\ 
{\tt\small zijiang@andrew.cmu.edu$^3$, zonghaiyao@umass.edu$^4$, hamidreza.kasaei@rug.nl$^2$, alex.li@ucl.ac.uk$^5$}
}

\maketitle
\ificcvfinal\thispagestyle{empty}\fi

\begin{abstract}
Videos depict the change of complex dynamical systems over time in the form of discrete image sequences. Generating controllable videos by learning the dynamical system is an important yet underexplored topic in the computer vision community. This paper presents a novel framework, \mbox{TiV-ODE}, to generate highly controllable videos from a static image and a text caption. Specifically, our framework leverages the ability of Neural Ordinary Differential Equations~(Neural ODEs) to represent complex dynamical systems as a set of nonlinear ordinary differential equations. The resulting framework is capable of generating videos with both desired dynamics and content. Experiments demonstrate the ability of the proposed method in generating highly controllable and visually consistent videos, and its capability of modeling dynamical systems. Overall, this work is a significant step towards developing advanced controllable video generation models that can handle complex and dynamic scenes.
\end{abstract}

\vspace{-5mm}
\section{Introduction}
Controllable video generation is an important research problem, as it enables the creation of highly specific and targeted video content. 
In contrast to traditional video generation methods~\cite{saito2017temporal, tulyakov2018mocogan, vondrick2016generating}, which produce uncontrollable or unpredictable results that may conflict with underlying physics, controllable video generation allows for precise manipulation of various aspects of the video, such as appearance and motion of objects within the video. This level of controlled video generation is crucial for a wide range of applications, such as creating user-specified effects and personalized video content. Additionally, controllable video generation could be used to generate synthetic data for training and research purposes in various fields such as robotics~\cite{yang2015robot, kapelyukh2022dall, dai2023learning} and self-driving cars~\cite{choi2021shared, chen2021geosim}. Overall, the ability to generate high-quality controllable videos efficiently has the potential to enhance the way we create, exploit, and interact with videos.

\begin{figure}[!t]
\centering
\includegraphics[trim=65mm 25mm 63mm 25mm,clip,width=0.95\linewidth]{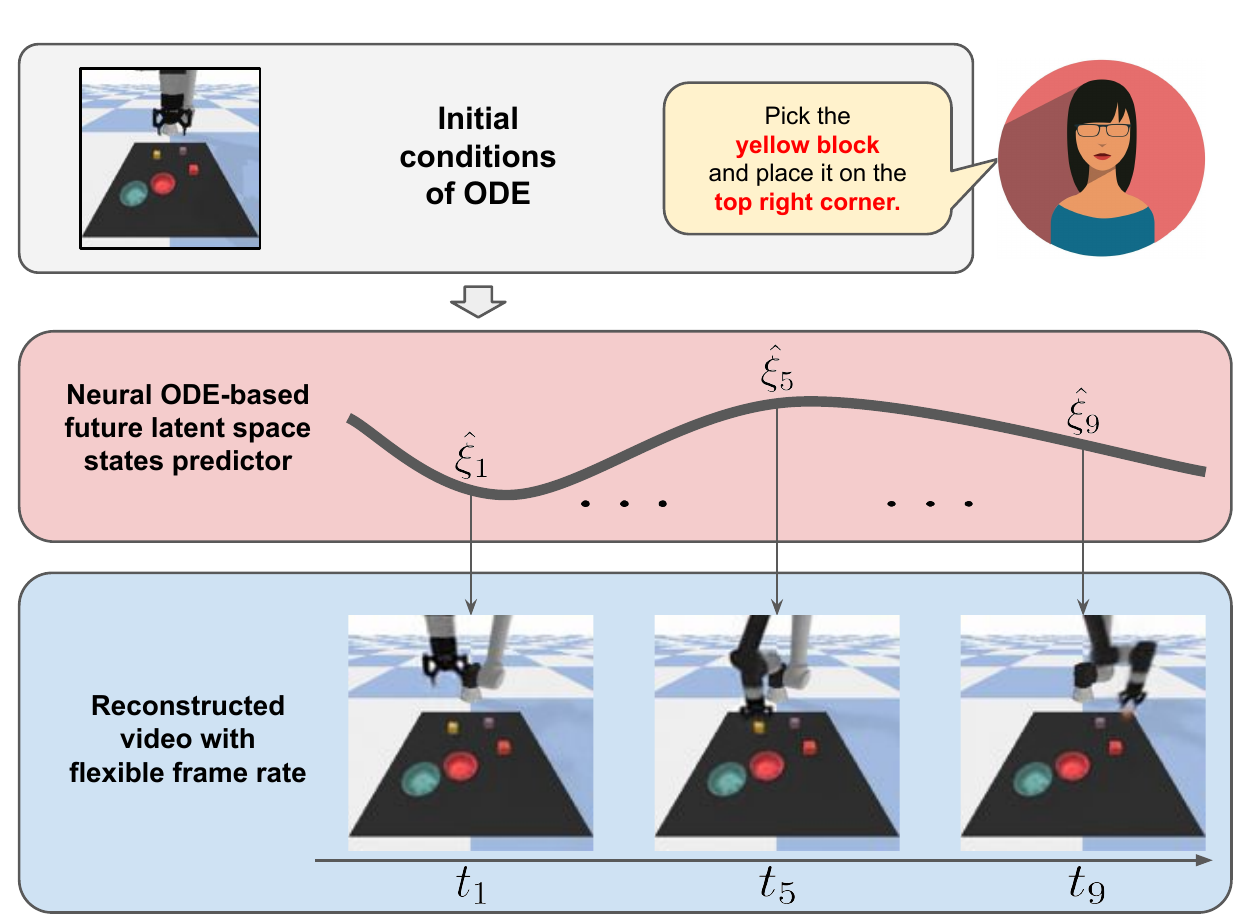}
\vspace{-0mm}
\caption{An example of controllable video generation from a static image and a text caption using our proposed TiV-ODE. The underlying dynamical system is modeled using Neural ODE. With the learned dynamical model and the given initial conditions, our model is capable of generating highly controllable and visually consistent video frames at any desired timesteps.} 
\vspace{-0mm}
\label{fig:imagine-basic}
\end{figure}

Controllable video generation aims to generate videos corresponding to given control signals. However, compared to static images, videos have an additional temporal dimension to be modeled. The appearance and states of objects in the video are tightly coupled with the temporal dimension -- the model must generate visually consistent content while predicting temporal changes based on motion cues to maintain motion consistency. While there are various video generation techniques, our work lies in the intersection between Image-to-Video and Text-to-Video methods. Image-to-Video methods~\cite{hao2018controllable, pan2019video, sheng2020high, yan2021videogpt} typically have no control over the motions in the generated videos, whereas Text-to-Video methods~\cite{mittal2017sync, wu2021godiva} offer limited control on the appearance of generated videos. Hence, to facilitate control over both motion and appearance in video generation, it is necessary to combine image and text signals.

Another vital limitation of previous controllable video generation methods is the lack of modeling of the underlying continuous dynamical system from videos. The dynamical system refers to a model that governs changes of the environment~(e.g. the dynamics of objects). In the predicted image sequences, the appearance and motion of observed objects shall always be consistent with given control signals -- consistency between the underlying dynamical system and the generated images. Prior methods typically model the underlying dynamical system as a discrete function of time, ignoring the fundamental difference between the continuous time dimension and the discrete image dimension as discussed in~\cite{park2021vid, blattmann2021understanding}. Such approaches limit the ability to generate videos with flexible frame rates and handle videos with arbitrary frame rates. For instance, given video frames at $t = 0s$ and $t = 5s$, existing methods cannot generate a video frame at $t=1.8s$ as they do not have a model for the underlying continuous dynamical system. Moreover, in various applications, such as slow-motion video processing~\cite{jin2019learning} or high-speed camera video processing~\cite{paliwal2020deep}, the regular timestep assumption does not hold. Therefore, a new controllable video generation method is needed which should be capable of generating highly controllable videos while correctly modeling the underlying continuous dynamical system.

To address these limitations, we developed a framework \textbf{T}ext-\textbf{i}mage-to-\textbf{V}ideo Ordinary Differential Equation~(\mbox{\textbf{TiV-ODE}}). Firstly, our proposed method leverages the advantages of both Image-to-Video methods and Text-to-Video methods since images and texts are two complementary signals, static images provide rich visual information, while text captions describe the dynamic processes within videos in human language. By combining image input and text input, both the visual appearance and the physical motions within the videos can be further constrained to allow high-level control of the video content. Secondly, stemming from the physical modeling of dynamical systems~\cite{chang2019antisymmetricrnn, strogatz2018nonlinear}, Neural ODE~\cite{chen2018neural} is incorporated in our proposed method to model the underlying continuous dynamical systems as ordinary differential equations~(ODEs). By solving the ODE at arbitrary timestamps, our model is able to generate videos with flexible frame rates efficiently~(See Figure~\ref{fig:imagine-basic}). To the best of our knowledge, the proposed method is a new approach to solving controllable video generation problems. We summarize our contributions as follows:

\begin{itemize}[leftmargin=*]
    \setlength{\itemsep}{1pt}
    \setlength{\parskip}{0pt}
    \setlength{\parsep}{0pt}

    \item We proposed a novel video generation framework, \mbox{TiV-ODE}, which is capable of generating highly controllable and visually consistent videos conditioned on a single image and a text caption.
    \item Our proposed method is able to generate videos with flexible frame rates by leveraging Neural ODE to model the underlying continuous dynamical system from videos.
    \item We created a new dataset, the Synthetic Robot Pick-and-Place dataset -- video sequences depicting a robot performing pick-and-place tasks with corresponding text captions -- for evaluating our method and demonstrating its effectiveness. We also performed experiments on existing datasets such as CATER and Moving MNIST and showed improvements compared to previous works. 
\end{itemize}

\section{Related Work}
In this section, we review the relevant controllable video generation methods and discuss the previous research that enhances video generation by understanding the underlying dynamical system from videos.

\subsection{Controllable Video Generation}
A synthetic video can be generated in a number of ways using various conversion techniques. The controllable video generation methods that are most pertinent to our work include Image-to-Video and Text-to-Video methods. It should be noted that unconditional video generation methods~\cite{saito2017temporal, tulyakov2018mocogan, vondrick2016generating} will not be discussed in this section as they are out of the scope of this paper.

\textbf{Image-to-Video} methods generate video sequences conditioned by given images. However, as a static image provides no motion clues, to facilitate video generation with editable scene dynamics, these methods require additional input to control the motions within the generated video, such as sparse trajectories~\cite{hao2018controllable, blattmann2021understanding}, and semantic masks~\cite{pan2019video, sheng2020high}. The method in~\cite{hao2018controllable} conditions video generation with a set of input optical flows. An auto-encoder network is used to generate video sequences from the given images and flow vectors. Similar to the method in~\cite{hao2018controllable}, the method in~\cite{blattmann2021understanding} also uses sparse trajectories to inject motion information during video generation. The methods in~\cite{pan2019video, sheng2020high} use semantic masks to inject motion information since objects from different categories have visually different motion patterns~(e.g. cars, buildings). Existing Image-to-Video methods can only achieve low-level control of the generated videos, thus they are not suitable to be used to generate videos with complex motions.

\textbf{Text-to-Video}
methods aim at generating video sequences from text captions. However, the appearance and motion information in the text caption is highly ambiguous leading to unavoidable uncertainties in generated videos. Sync-Draw~\cite{mittal2017sync} is the first framework proposed to solve Text-to-Video tasks. A region-of-interest VAE framework is proposed in~\cite{mittal2017sync} to generate video sequences that are consistent with the given simple text captions. Recently, GODIVA~\cite{wu2021godiva} was proposed to generate open-domain videos from given text captions in an auto-regressive way. A \mbox{VQ-VAE}~\cite{van2017neural} is used in~\cite{wu2021godiva} to represent the video content and generate visual tokens. Then, three attention models across temporal, column, and row dimensions, are used to generate temporally consistent videos conditioned by the caption. Given the ambiguous nature of the text, Text-to-Video methods can only achieve a low level of control over the generated videos. As a result, the appearance and motions within generated videos are mostly determined by the training dataset.

There is limited research work focused on combining the advantage of both \mbox{Image-to-Video} methods and \mbox{Text-to-Video} methods. To the best of our knowledge, the work in~\cite{hu2022make} is the closest one to our work. The work in~\cite{hu2022make} proposed a framework, MAGE, which generates videos from images with text captions. A motion embedding is used in MAGE~\cite{hu2022make} to memorize the motion patterns after observing the whole video, while our method formulates the underlying continuous dynamical system as an ordinary differential equation~(ODE) and approximates it using a neural network~\cite{chen2018neural}. Compared to MAGE, our method is able to generate videos with flexible frame rates, which greatly widens its potential applications. Moreover, during the inference phase, MAGE generates videos in an auto-regressive way while our method can generate all video frames simultaneously by directly solving the ODE at the desired timesteps, which is more efficient. A detailed comparison between our method and the MAGE are presented in Section~\ref{sec:exp}.

\subsection{Dynamical System Understanding from Videos}
\label{subsec:dyna}
Videos depict the dynamics of objects in the form of discrete image sequences. Modeling and understanding such dynamical systems from videos is important for video processing. Previous methods typically model the underlying dynamical system using an RNN-based structure~\cite{lin2020improving,kipf2021conditional, li2021object} or a transformer-based structure~\cite{wu2022slotformer} that can represent the temporal information. Another energy-based Spatial-Temporal generative model was proposed in~\cite{xie2017synthesizing, xie2019learning}, which learns the dynamic patterns in video sequences by matching the synthesized signals generated by sampling the Langevin dynamics to the observed training signals. Moreover, a set of dynamic latent variable models were proposed in \cite{xie2019learning, xie2020motion}.
The dynamic generator model in~\cite{xie2019learning} generates video sequences in a non-linear auto-regressive way. An alternating back-propagation through time algorithm was proposed in~\cite{xie2019learning} to train the dynamic latent variable model. The motion-based generator model in~\cite{xie2020motion} focuses on disentangling the appearance, trackable motions, and untrackable motions within the video. The model in~\cite{xie2020motion} is trained in an unsupervised way by directly using the Maximum likelihood learning algorithm without relying on any assisting module~(e.g. discriminator in GANs~\cite{goodfellow2020generative}). However, since these methods are mostly Video-to-Video methods, which are affected by the dynamics bias from the training data, they failed to generate videos with editable dynamics. To allow controllable video generation while modeling the underlying dynamical system, the method in~\cite{blattmann2021understanding} models the dynamical system as $n^{th}$ order ODEs which are approximated by using hierarchical RNNs. However, the method in~\cite{blattmann2021understanding} relies on the interplay between object parts and holds the assumption that the background remains static, thus it can only generate videos with simple motions.

\subsection{Neural ODE}
Neural Ordinary Differential Equations~(Neural ODEs, NODEs)~\cite{chen2018neural} interprets the forward pass of a ResNet~\cite{he2016deep} as solving an ordinary differential equation. It is designed to model the temporal evolution of any dynamical system. Recent works~\cite{rubanova2019latent, de2019gru, yildiz2019ode2vae} have shown the power of Neural ODE for modeling time series. Augmented Neural ODEs~(ANODEs)~\cite{chen2018neural} was proposed to extend the original Neural ODE by augmenting the latent space, which makes it a universal approximator~\cite{teshima2020universal, zhang2020approximation}. The method in~\cite{kanaa2021simple} introduces the Neural ODE into video generation tasks to model time-continuous dynamics within the videos over a continuous latent space. Vid-ODE~\cite{park2021vid} combines Neural ODE with GAN~\cite{goodfellow2020generative} to reconstruct video frames from the trajectory of latent dynamics, which allows high-quality future frame prediction and video interpolation.

\begin{figure*}[!t]
\centering
\includegraphics[trim=7mm 12mm 1mm 0mm,clip,width=0.95\linewidth]{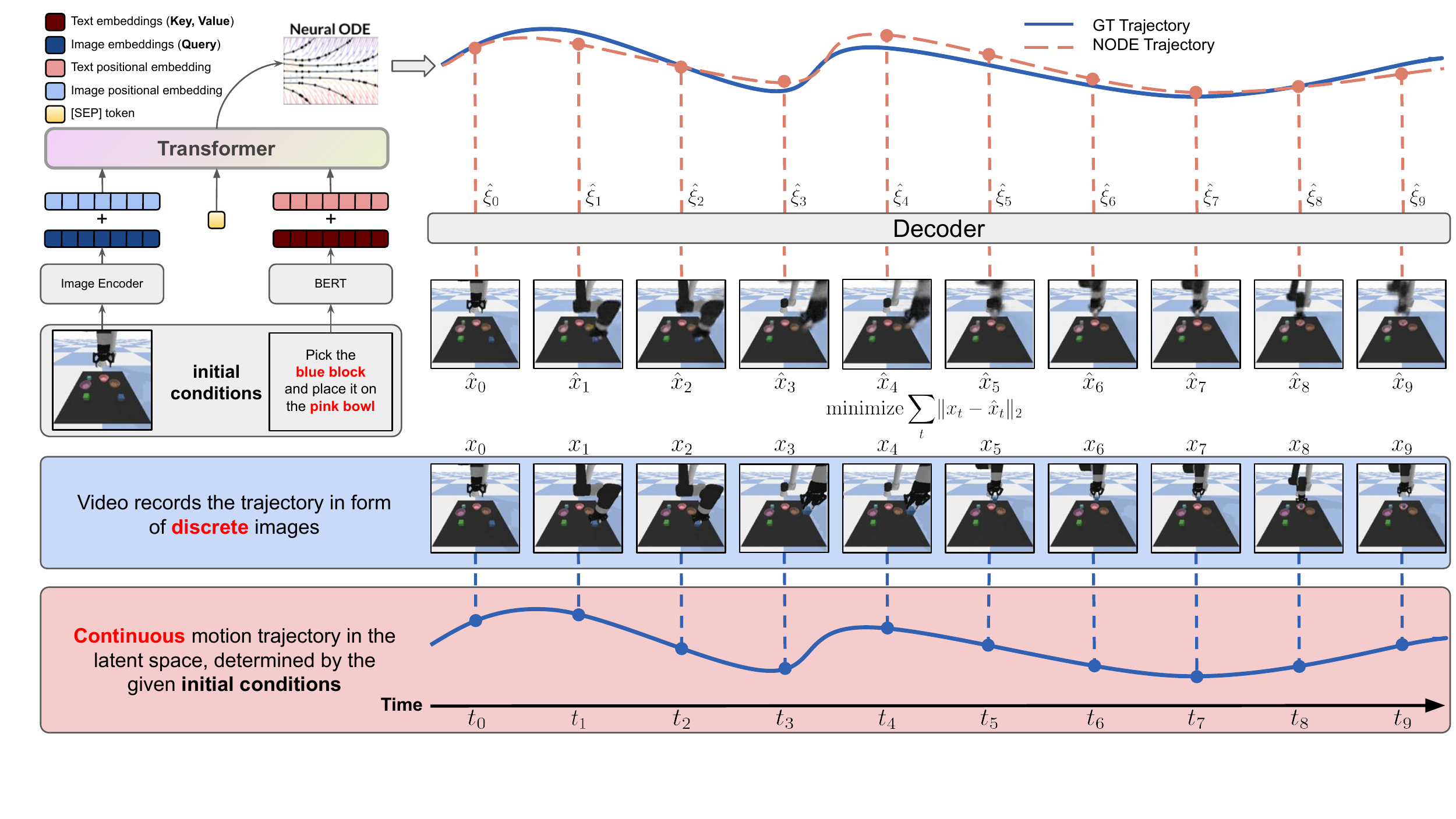}
\vspace{-0mm}
\caption{System structure of TiV-ODE. The continuous physical motions of objects are recorded as a set of discrete video frames. The latent vectors of the video are assumed to follow an ODE trajectory that corresponds to the motions of objects~(See the blue and red blocks). Given the initial image and the text caption, the whole video sequence is generated as follows:~(1) the input image and text are encoded by the image encoder and text encoder respectively.~(2) Together with positional embeddings, the image embeddings and text embeddings are fused by a transformer to generate a Text-image embedding.~(3) The Text-image embedding is used as the initial condition of Neural ODE, then the Neural ODE is solved at the desired timesteps using a numerical ODE solver to generate latent vectors at every timestep.~(4) The generated latent vectors are quantized by the codebook and decoded by the image decoder to generate video frames at every timestep. The training objective of our model is to minimize the distance between each pair of data points at each desired timestep.} 
\vspace{-0mm}
\label{fig:imagine}
\end{figure*}

\section{Text-image-to-Video ODE for Controllable Video Generation}
\label{sec:tech}
In this section, we first explain how we formulate the problem of controllable video generation by learning the dynamical system using Neural ODE, followed by a discussion on the general architecture of our proposed TiV-ODE. Then, details of our TiV-ODE, including the VQ-VAE for image generation, the text-image fusion module, and the Neural ODE module, will be presented individually.

\subsection{Problem Formulation}
\label{subsec:problem-formulation}
This paper targets the Text-image-to-Video task with modeling of the underlying continuous dynamical system. Let $x_t \in \mathcal{X} \subset \mathbb{R}^{N}$ be an image observation of the system (defined in the image sample space $\mathcal{X}$) at time point $t$, and $s \in S \subset  \mathbb{R}^{L}$ be the text caption (defined in the text sample space $S$). We aim to model the dynamical system defined over the text-image domain ($\mathcal{X} \times S$) such that, given a text caption $s$ and an image observation $x_0$ as the initial conditions, our model can generate the image observations $x_t$ for any $t \geq 0$. Unlike the previous method~\cite{hu2022make} which models a dynamical process using a discrete state-transition, i.e. \mbox{$x_{t} = \text{RNN}(x_{t-1}, s)$}, we model the system as a continuous vector field, $(x_0, s)~\mapsto~x(t),~\forall~t~\in (0,1)$, that is saying we want to approximate a function $x(t) = F(x_0, s, t), \forall t \in (0,1)$. The training objective of our proposed method is to approximate the continuous vector field by minimizing the distance between each data point and its prediction, i.e. video frame $x_t$ and the generated image $\hat{x_t}$.

\begin{figure*}[!ht]
\begin{subfigure}{\textwidth}
\centering
\includegraphics[trim=0mm 110mm 0mm 0mm,clip,width=0.9\textwidth]{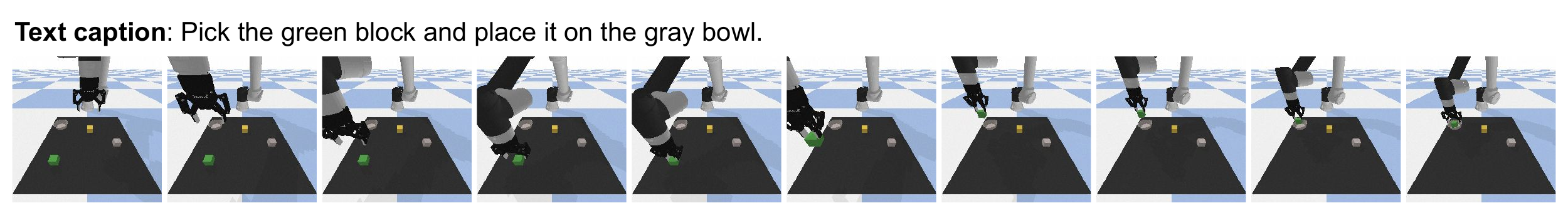}
\caption{Synthetic Robot Pick-and-Place}
\end{subfigure}
\begin{subfigure}{\textwidth}
\centering
\includegraphics[trim=0mm 110mm 0mm 0mm,clip,width=0.9\textwidth]{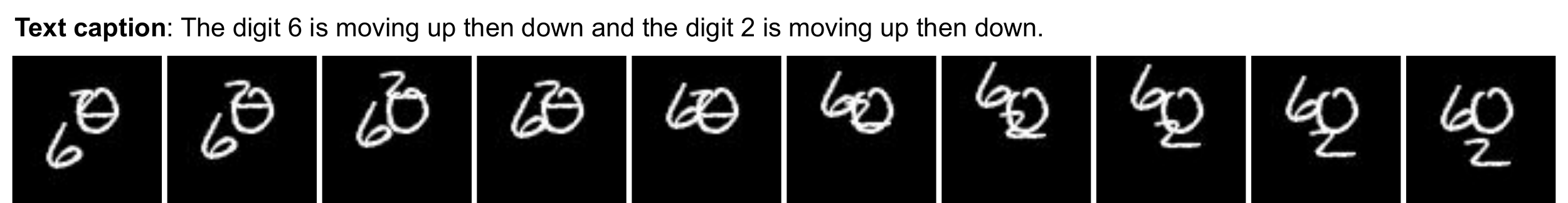}
\caption{Moving MNIST, triple digits}
\end{subfigure}
\begin{subfigure}{\textwidth}
\centering
\includegraphics[trim=0mm 110mm 0mm 0mm,clip,width=0.9\textwidth]{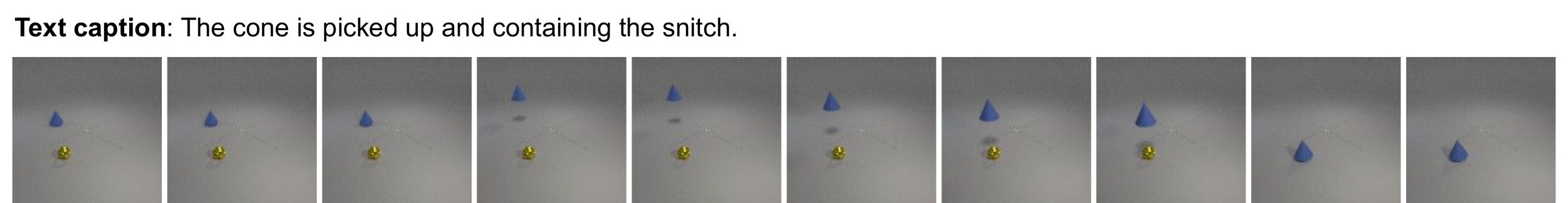}
\caption{CATER-v1}
\end{subfigure}
\begin{subfigure}{\textwidth}
\centering
\includegraphics[trim=0mm 105mm 0mm 0mm,clip,width=0.9\textwidth]{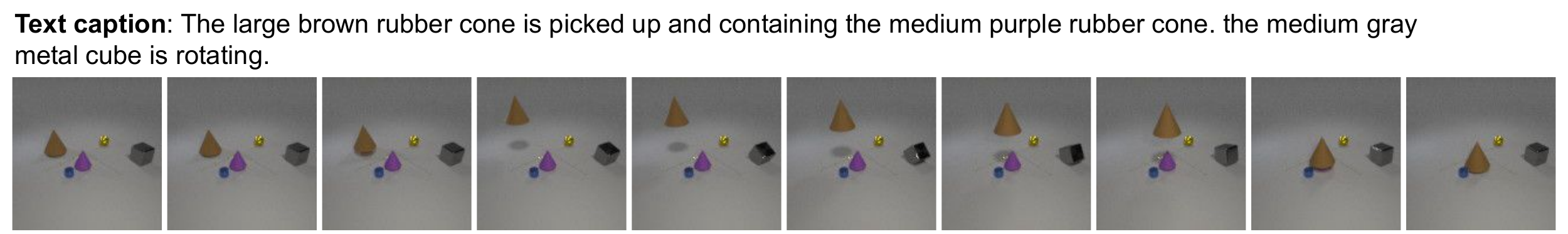}
\caption{CATER-v2}
\end{subfigure}
\vspace{-0mm}
\caption{Samples from synthetic robot pick-and-place dataset, moving MNIST dataset~\cite{mittal2017sync}, and CATER datasets~\cite{johnson2017clevr, girdhar2019cater}.}
\label{fig:samples}
\vspace{-0mm}
\end{figure*}

\subsection{Text-image-to-Video ODE}
\label{subsec:overview}
The overall architecture of the proposed method, TiV-ODE, is illustrated in Figure~\ref{fig:imagine}. Our approach uses the \mbox{VQ-VAE}~\cite{van2017neural} model for image generation. Compared to other encoder-decoder architectures, the \mbox{VQ-VAE} embeds high-dimensional visual images into a compact discrete codebook. Given the initial static image, $x_0$, and the text caption, $s$, the input image $x_0$ is encoded as a set of image embeddings by the \mbox{VQ-VAE} encoder, while the text caption $s$ is tokenized and encoded into a set of text embeddings using BERT~\cite{devlin2018bert}. After that, the image embeddings and text embeddings are aligned and fused using a multi-modal transformer~\cite{vaswani2017attention, arici2021mlim}. Image embeddings are used as \textbf{Query}, while text embeddings are used as \textbf{Key} and \textbf{Value}. The Text-image embeddings generated by the transformer are then used as the initial condition of the Neural ODE~\cite{chen2018neural}. Afterward, using this initial condition, the Neural ODE module learns the dynamical system behind the videos by approximating the continuous vector field during the training phase. Hence, the latent vector for any time point $t$ can be generated by solving the Neural ODE at time $t$. The generated latent vector is then quantized by the codebook and decoded by the \mbox{VQ-VAE} decoder to generate a video frame $\hat{x_t}$ at time $t$.

\subsection{VQ-VAE for Image Generation}
\label{subsec:VQ-VAE}
The \mbox{VQ-VAE}-based encode-decoder structure~\cite{van2017neural} is used in our proposed method for image generation. It is important to note that before training our TiV-ODE, the \mbox{VQ-VAE}~\cite{van2017neural} module is pre-trained separately on each dataset and then fine-tuned to make the codebook more suitable for representing the video frames. A typical \mbox{VQ-VAE} model is composed of an encoder $E$, a decoder $D$, and a discrete codebook $\mathcal{E} \in \mathbb{R}^{K \times N}$, which is basically a list of vectors $\{e_0, e_1, \dots, e_K\}$, where $K$ is the size of the codebook and $N$ is the dimension of the codebook. The encoder encodes input image $x \in \mathbb{R}^{H \times W \times 3}$ into a latent vector $z_e(x) = E(x) \in \mathbb{R}^{h \times w \times c}$, where $h=H/n, w=W/n$, $n$ is the downsampling ratio of the encoder, and $c$ is the output dimension of the encoder. Then the latent vector $z_e(x)$ is compared to all vectors in the codebook, and the closest codebook vector~(euclidean distance) is input into the decoder to generate the reconstructed image. Mathematically, this is written as $\hat{x} = D(z_q(x))$, where $z_q(x) = e_k, k = \operatorname*{argmin}_i \lVert z_e(x) - e_i \rVert_2$. The training objective of \mbox{VQ-VAE} is to minimize:
\begin{equation}
\resizebox{0.9\hsize}{!}{
$\log(p(x|z_q(x))) + \lVert \operatorname*{sg}[z_e(x)] - e \rVert_2^2 + \beta \lVert z_e(x) - \operatorname*{sg}[e] \rVert_2^2,$}
\end{equation}
where $\operatorname*{sg}[*]$ stands for the stop gradient operation. The first term is the standard reconstruction loss. The second and third terms are the codebook alignment loss to make the selected codebook vector $e_k$ close to the latent vector $z_e(x)$ by updating the codebook and encoder respectively, $\beta$ is the commit loss weight. The \mbox{VQ-VAE} in our method is trained using the Expectation Maximization (EM) algorithm~\cite{roy2018theory}.

\subsection{Text-Image Fusion Module}
\label{subsec:fusion}
Inspired by the MLIM proposed in~\cite{arici2021mlim}, a multi-modal transformer~\cite{vaswani2017attention} is used in our TiV-ODE to fuse the input image and the text caption. Specifically, the image embedder is the encoder of a pre-trained \mbox{VQ-VAE}. The 2D positional embedding, similar to the one in~\cite{vaswani2017attention, wang2021translating}, is added to each image token to keep the positional information. The text caption is firstly tokenized by the BERT's~\cite{devlin2018bert} tokenizer, then, the text embeddings are obtained from BERT’s~\cite{devlin2018bert} word embeddings. The positional embeddings for text tokens come with the BERT’s~\cite{devlin2018bert} word embeddings. During the multi-modal transformer operation, the image embeddings are used as \textbf{Query}, while the text embeddings are used as \textbf{Key} and \textbf{Value}.

\subsection{Neural ODE for Modeling Dynamical System}
\label{subsec:node}
Neural ODE is the essential part of our proposed \mbox{TiV-ODE}, which models the underlying dynamical system in the latent space as a continuous ordinary differential equation. To better represent the complex dynamical systems~(e.g. the trajectories of moving objects are overlapped with each other), we adopted the augmented Neural ODE~\cite{dupont2019augmented} instead of the original one~\cite{chen2018neural}.

Let $\xi_t$ be the state of the dynamical system in the latent space at an arbitrary time $t$, and let $f$ be the ordinary differential equation that describes the dynamical system. The differential function $f$ is approximated by an estimator $f_{\theta} \simeq f$ parameterized by $\theta$. A time-dependent convolutional network is used in our method as the $f_\theta$. Then, the dynamical system modeled by the Neural ODE satisfies a Cauchy problem,
\begin{equation}
\frac{\partial\xi(t)}{\partial t} = f_{\theta}(\xi(t), t), \;\;  \xi_0 = \operatorname*{Transformer}(z_e(0), s).
\end{equation}
Thus, the state of the dynamical system can be obtained at any timestep by invoking an ODE solver~(e.g. Runge-Kutta of Dormand-Prince~\cite{dormand1980family} in our setting) to compute a numerical approximation of the integral of the dynamical system from the initial value:
\begin{equation}
\begin{split}
\xi(t_i) = \operatorname*{ODESolver}(f_{\theta}, \xi_0, (t_0, t_i)) \simeq \\ 
\xi_0 + \int_{t_0}^{t_i} f(\xi(\tau), s, \tau)d\tau = \xi_{i}.
\end{split}
\label{eq:odesolver}
\end{equation}
Then, the state of the dynamical system in latent space at timestep $t$, $\xi(t)$, is quantized by the codebook and decoded by the \mbox{VQ-VAE} decoder to reconstruct the video frame at timestep $t$, $\hat{x}_t$.

\section{Experiments}
\label{sec:exp}
In this section, we first introduce the datasets used to evaluate our method, including our proposed synthetic robot pick-and-place dataset, Modified Moving MNIST dataset~\cite{mittal2017sync, hu2022make}, and the CATER dataset~\cite{johnson2017clevr, girdhar2019cater}. Then we present the quantitive results on these datasets and compare our method with MAGE~\cite{hu2022make}. After that, we demonstrate the controllability of video generation~(See Section~\ref{subsec:exp-controll}) and the ability to model continuous dynamical systems~(See Section~\ref{subsec:exp-dynamic}) of our method by presenting videos generated with different text captions, and videos with different frame rates. Finally, our ablation studies validated the effectiveness of our model designs. For a better understanding of the videos generated by our model, more details are available in the supplementary video accompanying this paper.

\subsection{Datasets}
\label{subsec:datasets}
\textbf{Modified Moving MNIST dataset}~\cite{mittal2017sync, hu2022make}. Instead of the original moving MNIST datasets~\cite{mittal2017sync}, we used a modified version introduced in~\cite{hu2022make}. Five motion patterns are included in moving MNIST datasets, up then down, left then right, down then up, right then left, and static. We use three types of moving MNIST datasets to evaluate our method, single moving MNIST dataset, double moving MNIST dataset, and triple moving MNIST dataset.

\textbf{CATER datasets}~\cite{johnson2017clevr, girdhar2019cater} were introduced in~\cite{girdhar2019cater} based on the CLEVR dataset~\cite{johnson2017clevr}. There are four different motion patterns in the dataset, ``contain", ``slide", ``rotate", and ``pick-place". Each video in the dataset contains one or two random actions. We follow the same settings used in~\cite{hu2022make} to generate CATER-v1 dataset and CATER-v2 dataset. The CATER-v1 dataset contains scenes with 2 objects and one random motion. The CATER-v2 dataset contains scenes with 3 to 6 objects with two random motions.

\textbf{Synthetic Robot Pick-and-Place dataset}. We propose the synthetic robot pick-and-place dataset based on the simulation environment used in~\cite{zeng2022socraticmodels, saycan2022arxiv}. Each sample in this dataset contains a video sequence depicting a robot pick-and-place process and a text caption specifying the pick-up and placement targets. We constructed this dataset and used it to evaluate our model, showing that our method is capable of generating videos depicting intricate robotics processes. Our results highlight the potential of our model for future robotics research.

Samples from each dataset are depicted in Figure~\ref{fig:samples}. Due to the limited space, details of these datasets are presented in the supplementary material.

\begin{figure*}[!t]
\begin{subfigure}{\textwidth}
\centering
\includegraphics[trim=0mm 100mm 0mm 3mm,clip,width=0.9\textwidth]{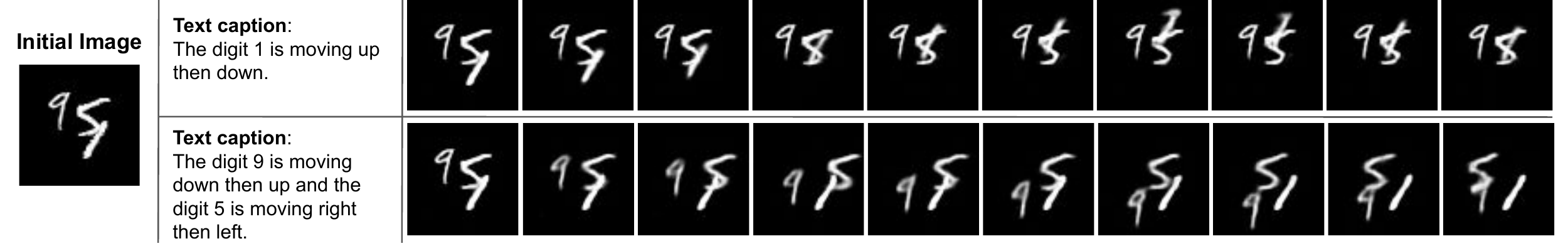}
\caption{Moving MNIST, triple digits}
\label{subfig:mnist-results}
\end{subfigure}
\begin{subfigure}{\textwidth}
\centering
\includegraphics[trim=0mm 100mm 0mm 3mm,clip,width=0.9\textwidth]{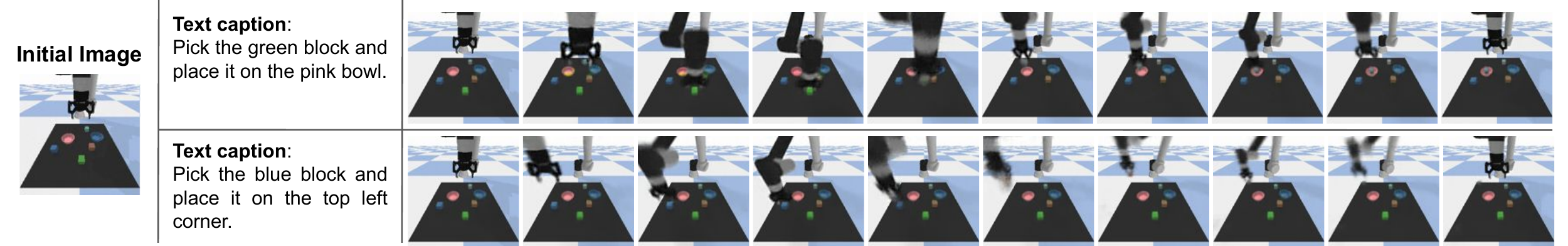}
\caption{Synthetic Robot pick-and-place}
\label{subfig:robot-results}
\end{subfigure}
\begin{subfigure}{\textwidth}
\centering
\includegraphics[trim=0mm 77mm 0mm 0mm,clip,width=0.9\textwidth]{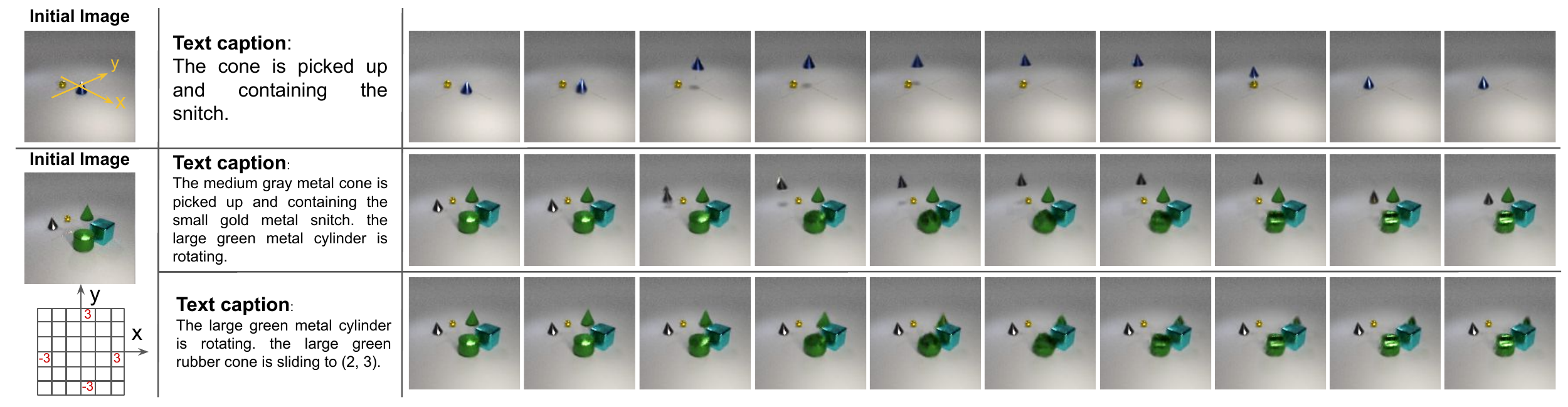}
\caption{CATER-v1 and CATER-v2}
\label{subfig:cater-results}
\end{subfigure}
\vspace{-0mm}
\caption{Results of controllable video generation on robot moving MNIST dataset~\cite{mittal2017sync}, synthetic pick-and-place dataset, and CATER datasets~\cite{johnson2017clevr, girdhar2019cater}. The coordinate system used in CATER datasets is demonstrated in~(c). It is noteworthy that with the fixed initial image, our TiV-ODE can precisely manipulate different objects specified by different text captions and generate videos with both visual consistency and motion consistency.}
\label{fig:controllable-video-generation}
\vspace{-0mm}
\end{figure*}

\subsection{Quantitive Results}
\label{subsec:quant}
This section presents the quantitive results of our method on the datasets mentioned in Section~\ref{subsec:datasets}. Furthermore, we compare our approach with MAGE~\cite{hu2022make} which is the current state-of-the-art in the Text-image-to-Video domain. Training and testing protocols are presented in the supplementary material.

The quantitive results, including SSIM~\cite{wang2004image} and PSNR~\cite{huynh2008scope, hore2010image}, on the moving MNIST datasets are reported in Table~\ref{tab:mnist}. By comparing all results, our method slightly outperforms the MAGE~\cite{hu2022make} in terms of SSIM and performs competitively in terms of PSNR. The lower performance on PSNR is due to the approximation of the ODE solvers. Our model generates video frames by solving the learned ODE at desired timesteps using a numerical ODE solver as presented in Equation~\ref{eq:odesolver}. Thus, there may exist residual shadows in the generated video frames~(See the digit 1 and 5 in Figure~\ref{subfig:mnist-results}) which leads to lower PSNR values, and, in some cases, lower SSIM values.

The quantitive results on the CATER datasets and synthetic robot pick-and-place dataset, including both pixel-based metric~(SSIM~\cite{wang2004image}) and perceptual metrics~(image-level Fréchet inception distance~(FID)~\cite{heusel2017gans}, and learned perceptual image patch similarity~(LPIPS)~\cite{dosovitskiy2016generating}), are reported in Table~\ref{tab:cater} and Table~\ref{tab:robot}, respectively. The quantitive results show that our method outperforms MAGE~\cite{hu2022make} in terms of FID, and LPIPS while performing competitively on the SSIM. Additionally, we present the inference speed of TiV-ODE and MAGE, referring to the time required to generate a batch of videos, with a batch size of 16 and a video length of 10 frames. Our TiV-ODE is significantly more efficient than the MAGE~(See Table~\ref{tab:cater}), since our approach does not rely on an auto-regressive methodology.

\begin{table}[!t]
\centering
\caption{Quantitive comparison of our proposed TiV-ODE and MAGE~\cite{hu2022make} on moving MNIST datasets~\cite{mittal2017sync}. The quantitative results of MAGE are taken from~\cite{hu2022make}.}
\resizebox{\linewidth}{!}{
\begin{tabular}{c|c|cc}
\hline
Datasets                                                                                & Methods       & SSIM~$\uparrow$ & PSNR~$\uparrow$  \\ \hline
\multirow{2}{*}{\begin{tabular}[c]{@{}c@{}}Single moving\\ MNIST\end{tabular}}          & MAGE~\cite{hu2022make}           & 0.97 & \textbf{33.89} \\ \cline{2-4} 
                                                                                        & \textbf{TiV-ODE~(Ours)} & 0.97 & 31.8  \\ \hhline{====}
\multirow{2}{*}{\begin{tabular}[c]{@{}c@{}}Double moving\\ MNIST\end{tabular}}          & MAGE~\cite{hu2022make}           & 0.87 & \textbf{24.66} \\ \cline{2-4} 
                                                                                        & \textbf{TiV-ODE~(Ours)} & \textbf{0.90} & 23.95 \\ \hhline{====}
\multirow{2}{*}{\begin{tabular}[c]{@{}c@{}}Modified double\\ moving MNIST\end{tabular}} & MAGE~\cite{hu2022make}           & 0.85 & \textbf{23.24} \\ \cline{2-4} 
                                                                                        & \textbf{TiV-ODE~(Ours)} & 0.85 & 21.41 \\ \hline
\end{tabular}}
\label{tab:mnist}
\end{table}

\begin{table}[!t]
\centering
\caption{Quantitive comparison of our proposed TiV-ODE and MAGE~\cite{hu2022make} on CATER-v1 dataset and CATER-v2 dataset~\cite{mittal2017sync}. The quantitative results of MAGE are taken from~\cite{hu2022make}.}
\resizebox{\linewidth}{!}{
\begin{tabular}{c|c|cccc}
\hline
Datasets                      & Methods       & SSIM~$\uparrow$ & FID~$\downarrow$   & LPIPS~$\downarrow$ & \begin{tabular}[c]{@{}c@{}}Inference\\ speed\end{tabular}~$\downarrow$ \\ \hline
\multirow{2}{*}{CATER-GEN-v1} & MAGE~\cite{hu2022make}          & 0.96 & 62.66 & 0.20  & 0.8s                                                      \\ \cline{2-6} 
                              & \textbf{TiV-ODE~(Ours)} & 0.96 & \textbf{11.98} & \textbf{0.12}  & \textbf{0.06s}                                                     \\ \hhline{======}
\multirow{2}{*}{CATER-GEN-v2} & MAGE~\cite{hu2022make}          & \textbf{0.95} & 39.56 & 0.20  & 0.8s                                                      \\ \cline{2-6} 
                              & \textbf{TiV-ODE~(Ours)} & 0.93 & \textbf{38.12} & \textbf{0.18}  & \textbf{0.06s}                                                     \\ \hline
\end{tabular}}
\label{tab:cater}
\end{table}

\begin{table}[!t]
\centering
\caption{Quantitive comparison of our proposed TiV-ODE and MAGE~\cite{hu2022make} on synthetic robot pick-and-place dataset~\cite{mittal2017sync}. The quantitative results of MAGE are produced using the official implementation.}
\resizebox{\linewidth}{!}{
\begin{tabular}{c|c|ccc}
\hline
Datasets                                                                                & Methods       & SSIM~$\uparrow$ & FID~$\downarrow$   & LPIPS~$\downarrow$ \\ \hline
\multirow{2}{*}{\begin{tabular}[c]{@{}c@{}}Robot Pick-and-Place\\ dataset\end{tabular}} & MAGE~\cite{hu2022make}          & \textbf{0.94} & 33.69 & 0.18  \\ \cline{2-5} 
                                                                                        & \textbf{TiV-ODE~(Ours)} & 0.93 & \textbf{27.48} & \textbf{0.12}  \\ \hline
\end{tabular}}
\label{tab:robot}
\end{table}

\begin{figure*}[!t]
	\centering
	\includegraphics[trim=0mm 92mm 1mm 0mm,clip,width=\linewidth]{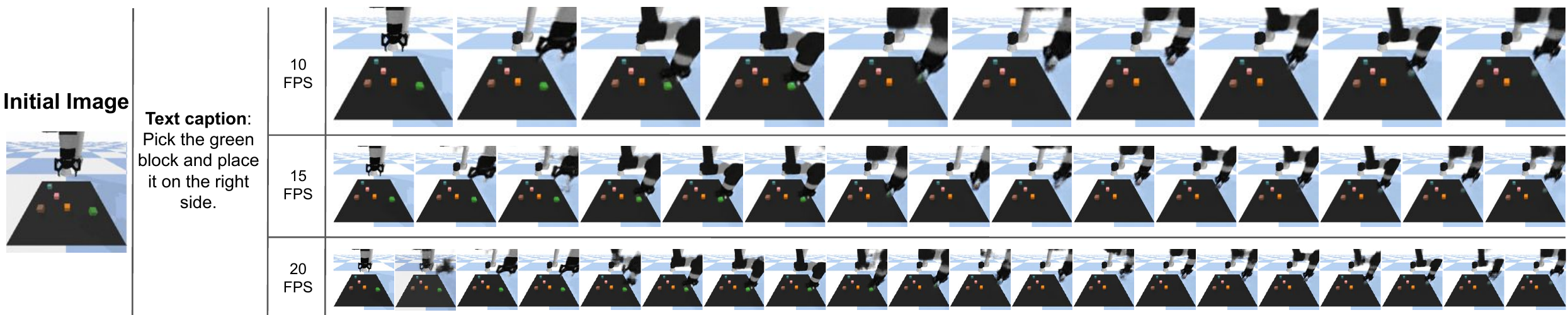}
	\vspace{-0mm}
	\caption{Generated videos with arbitrary frame rates by solving Neural ODE with different time intervals. Each row shows the snapshots of videos of 10 FPS, 15 FPS, and 20 FPS respectively. It is noteworthy that the same video sequence is generated, but more details were captured as the frame rate increases.} 
	\label{fig:frame-rates}
\end{figure*}

\begin{figure*}[!t]
	\centering
	\includegraphics[trim=33mm 96mm 31mm 10mm,clip,width=\linewidth]{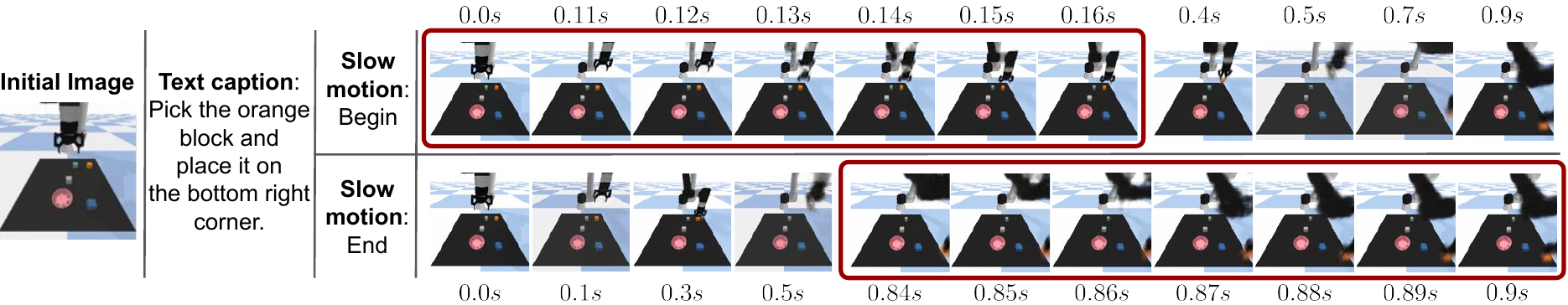}
	\vspace{-0mm}
	\caption{Generated videos with additionally created slow-motion effects which are generated by solving the Neural ODE using a set of denser timesteps at the desired slow-motion segment. The slow-motion frames in the figure are highlighted by the red blocks, which capture more detailed slow changes compared to the normal speed segments. } 
	\vspace{-0mm}
	\label{fig:slow-motion}
\end{figure*}

\subsection{Controllable Video Generation}
\label{subsec:exp-controll}
In this section, we present the test results on the moving MNIST dataset, synthetic robot pick-and-place dataset, and CATER datasets. The generated video sequences are shown in Figure~\ref{fig:controllable-video-generation}. We experimented with manipulating different digits on the Moving MNIST dataset (see Figure~\ref{subfig:mnist-results}) by using various text captions. Our results demonstrate that our model successfully recognizes and localizes the target digits specified by the text caption, producing video sequences with correct motions. On the synthetic robot pick-and-place dataset~(See Figure~\ref{subfig:robot-results}), we specify different sets of pick-ups and placement targets. The results show that our model can successfully establish a connection between the position in the image and the position described by the text~(e.g. the top left corner). On the CATER datasets~(See Figure~\ref{subfig:cater-results}), the results show that our method can successfully distinguish and localize objects with similar properties~(e.g. large green rubber cone, medium gray metal cone). Overall, these results show that our method yields promising performance in achieving highly controllable video generation with a given static image and a text caption.

\subsection{Video Generation with Different Frame Rates}
\label{subsec:exp-dynamic}
In this section, we demonstrate the ability of our method to model the underlying continuous dynamical system by showing the results of two experiments:~(\textit{i})~video generation with arbitrary frame rates;~(\textit{ii})~video generation with manually added slow motion effect.

\textbf{Video generation with arbitrary frame rates}. Our model is able to generate video sequences with arbitrary frame rates by solving the learned Neural ODE with different time intervals. Here we present the results from the synthetic robot pick-and-place dataset. Three video sequences with 10, 15, and 20~FPS are generated~(See Figure~\ref{fig:frame-rates}). 

\textbf{Video generation with slow-motion effects}. Our model is able to generate video sequences with manually added slow-motion effects that can be adjusted by using denser timesteps at the desired slow-motion segment. Formally, this effect is referred to as frame rate ramping. Here we present two generated video sequences from the synthetic robot pick-and-place dataset. One has slow motion at the beginning and the other has it at the end~(See Figure~\ref{fig:slow-motion}).

These results show that our model is capable of modeling the underlying continuous dynamical system from videos, and with the learned continuous dynamical system, our model is able to generate video sequences with arbitrary frame rates and manually added slow-motion effects~(\mbox{non-uniform} frame rates).

\subsection{Ablation Study}

\label{subsec:ablation}
We have conducted an ablation study on the CATER-v1 dataset to justify the core design of our TiV-ODE, the Neural ODE module. We replace the Neural ODE module with a step-wise transition module, i.e. the transition model receives the video frame at $t$ and predicts the frame at $t+1$,  to generate the video sequence. The results are presented in Table~\ref{tab:ablation}. 
\begin{table}[!t]
\centering
\caption{Quantitive results of the ablation study.}
\vspace{-2mm}
\resizebox{0.8\linewidth}{!}{
\begin{tabular}{c|ccc}
\hline
           & SSIM$\uparrow$ & FID$\downarrow$   & LPIPS$\downarrow$ \\ \hline
TiV-TransAll  & 0.90 & 85.2 & 0.31  \\ \hline
TiV-TransNext & 0.91 &  70.1 & 0.28  \\ \hhline{====}
\textbf{TiV-ODE}    & \textbf{0.96} & \textbf{11.98} & \textbf{0.12}  \\ \hline
\end{tabular}}
\label{tab:ablation}
\vspace{-2mm}
\end{table}
\textbf{TiV-TransAll} denotes that the transition model only receives the first frame, and the following video frames are generated in an iterative way. \textbf{TiV-TransNext} denotes that the transition model observes the whole video sequence during the training phase, while in the testing phase, the video sequences are generated in an iterative way. The results of the ablation study support the effectiveness of our model design. The ODE settings in our model enable it able to correctly model the underlying dynamical system by learning to generate videos from the initial conditions.

\section{Conclusion}
In this paper, we present a novel controllable video generation model which is able to generate highly controllable videos conditioned by a static image and a text caption. Moreover, our framework models and learns the underlying continuous dynamical system using Neural ODE. To show the potential of our model in robotics research, we created a new robot pick-and-place dataset to evaluate our model, as well as using the existing moving MNIST datasets and CATER datasets. Experiments results showed that our method yields promising results in terms of controllable video generation and dynamical system modeling. This work moves a significant step towards solving the challenging controllable video generation task and has the potential for downstream applications in robotics.

\textbf{Limitations and future works}. Firstly, training and solving the Neural ODE is time-consuming, especially when the motion patterns in the videos are complex, thus longer training time is required. Moreover, our model only ``sees" the first frame which provides relatively weaker constraints on later video frames, and the whole video sequence is generated by numerically solving the learned ODE function and decoding the output of the ODE, which may result in blurred images. Future work aims to extend and improve the framework for more complex motion patterns and higher-quality videos.

{\small
\bibliographystyle{ieeetr}
\bibliography{manuscript}
}

\end{document}